\documentclass[conference]{IEEEtran}
\IEEEoverridecommandlockouts
\usepackage{epsfig}
\usepackage{graphicx,subcaption}
\usepackage{epstopdf}
\usepackage[justification=centering]{caption}
\usepackage{algorithm,algorithmic}
\usepackage{array}
\usepackage{makecell}
\usepackage{url}
\usepackage{commath}
\usepackage{amsmath,amsfonts,amscd,amssymb,latexsym,bm}
\usepackage{gensymb}      
\usepackage{tikz}
\usetikzlibrary{positioning, shapes.geometric}
\usepackage{cite}
\usepackage{bm}
\usepackage{amsmath,amssymb,amsfonts}
\usepackage{algorithm}
\usepackage{algorithmic}
\usepackage{graphicx}
\usepackage{textcomp}
\usepackage{xcolor}
\usepackage{array} 
\usepackage[justification=centering]{caption}
\usepackage{makecell}
\def\BibTeX{{\rm B\kern-.05em{\sc i\kern-.025em b}\kern-.08em
    T\kern-.1667em\lower.7ex\hbox{E}\kern-.125emX}}
\begin{document}

\title{Using Collision Momentum in Deep Reinforcement Learning based Adversarial Pedestrian Modeling \\
\thanks{This work was supported in part by the United States Department of Transportation under Award Number 69A3551747111 for the Mobility21 University Transportation Center. Any opinions, findings, conclusions, or recommendations expressed herein are those of the authors and do not necessarily reflect the views of the United States Department of Transportation.}
}

\author{\IEEEauthorblockN{Dianwei Chen}
\IEEEauthorblockA{\textit{Department of Electrical and } \\
\textit{Computer Engineering} \\
\textit{The Ohio State University}\\
Columbus, Ohio \\
chen.11181@osu.edu}
\and
\IEEEauthorblockN{Ekim Yurtsever}
\IEEEauthorblockA{\textit{Department of Electrical and } \\
\textit{Computer Engineering} \\
\textit{The Ohio State University}\\
Columbus, Ohio \\
yurtsever.2@osu.edu}
\and
\IEEEauthorblockN{Keith A. Redmill}
\IEEEauthorblockA{\textit{Department of Electrical and } \\
\textit{Computer Engineering} \\
\textit{The Ohio State University}\\
Columbus, Ohio \\
redmill.1@osu.edu} 

\and
\IEEEauthorblockN{Ümit Özgüner}
\IEEEauthorblockA{\textit{Department of Electrical and } \\
\textit{Computer Engineering} \\
\textit{The Ohio State University}\\
Columbus, Ohio \\
ozguner.1@osu.edu}
}
\maketitle

\begin{abstract}

Recent research in pedestrian simulation often aims to develop realistic behaviors in various situations, but it is challenging for existing algorithms to generate behaviors that identify weaknesses in automated vehicles' performance in extreme and unlikely scenarios and edge cases. To address this, specialized pedestrian behavior algorithms are needed. Current research focuses on realistic trajectories using social force models and reinforcement learning based models. However, we propose a reinforcement learning algorithm that specifically targets collisions and better uncovers unique failure modes of automated vehicle controllers. Our algorithm is efficient and generates more severe collisions, allowing for the identification and correction of weaknesses in autonomous driving algorithms in complex and varied scenarios.
\end{abstract}

\section{Introduction}
Automated Driving Systems (ADS) are becoming more popular and Advanced Driver Assistance Systems (ADAS) are being applied to commercially available vehicles \cite{chan2017advancements,bengler2014three,bar2014recent}. However, there are still several challenges that need to be solved in realistic road and traffic conditions \cite{bonnefon2016social} involving high-dimensional state spaces, complex decision environments, and complicated and computationally demanding processing and learning algorithms\cite{beringhoff2022thirty}. Another major challenge are edge-cases: unusual, infrequent, or abnormal situations and  interactions of traffic participants including pedestrians and other vulnerable road users \cite{helle2016testing}.

The unique weaknesses of Automated Vehicle (AV) stacks can be found with smart adversarial agents that can generate system failures and help to identify the failure modes \cite{mullins2017automated}. Currently, many methodologies are used to measure and model pedestrian behavior\cite{yang2018social, predhumeau2022agent, Bonneaud2012ABD, chen2017socially}. However, measures of accident lethality, including collision momentum, have not been considered in modeling pedestrians' intention using a reinforcement learning Markov decision-process framework.

This work proposes a framework algorithm based on the reinforcement learning (RL) method which considers vehicle and pedestrian accident lethality modeled by collision momentum in the reward functions. The experiments show that our work can explore the more dangerous edge-case scenarios and help the pedestrian agent to learn to explore the weaknesses and test the safety threshold of the AV driving algorithms. 
The RL pedestrian agent was utilized to identify a distinct failure mode in which the pedestrian seeks to increase collision lethality through the optimization of their trajectory.

The main contributions of this work are:
\begin{itemize}
    \item A novel reward function involving collision-momentum is used to train a model to generate adversarial pedestrian behavior.
    \item A smarter pedestrian trajectory that can cause increased lethality compared to other pedestrian motion algorithms is generated.
    \item Through simulation studies, we show that our reinforcement learning algorithm based on Deep Deterministic Policy Gradient (DDPG) is a better choice, in comparison to other tested methods, for exploring the more dangerous edge-case scenarios and collisions.
\end{itemize}

\begin{figure}[!]
\centerline{\includegraphics[width=0.5\textwidth]{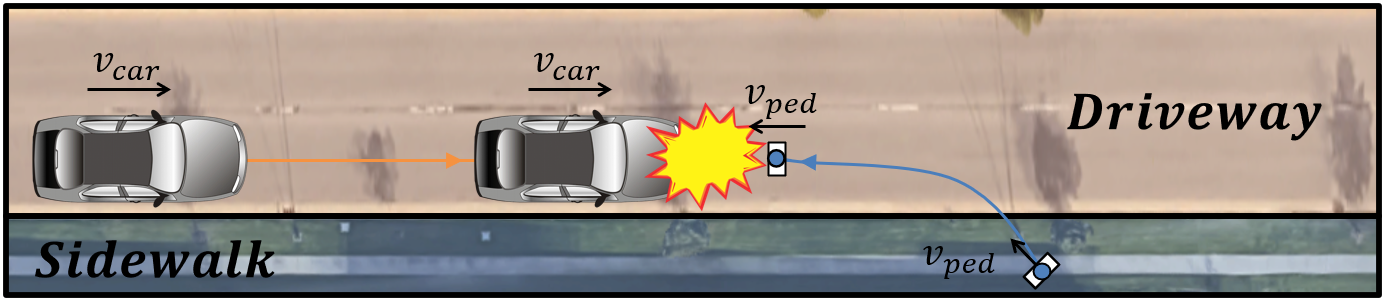}}
\caption{Proposed adversarial pedestrian scenario}
\label{ColPic}
\end{figure}

\section{Related Work}

\subsection{Adversarial pedestrian modeling}
Modeling pedestrian behavior is important because pedestrian accidents involving motor vehicles are a major concern. In 2020, over 7000 pedestrians were killed in the US \cite{noauthor_pedestrian_2022} and similar numbers are reported in other countries \cite{noauthor_reported_nodate}. With the increasing deployment of AVs \cite{wang2020safety}, it is necessary to test the limits of AV capabilities and improve safety \cite{liu2019edge}.


Pedestrian behavior modeling is a common approach that involves the combination of micro-simulation and macro-simulation techniques \cite{ni2011modelling}, \cite{martinez2017modeling} to model the behavior of both individual pedestrians and groups \cite{moussaid2010walking}. Machine learning techniques and data collection are also commonly used \cite{kielar2020artificial}. Many modeling methods are based on physics such as force and fluid dynamics, with Helbing and Molnár's social force model \cite{helbing1995social} being a popular foundation for later physics-based models. There is also a substantial body of research pertaining to agents \cite{bonneaud2012behavioral} modeled with a set of prescribed behavioral rules. Recently, there has been an increase in research utilizing deep networks and reinforcement learning as a method for modeling pedestrian behavior \cite{kiran2021deep}, which has gained widespread acceptance within the field. Most literature and research focuses on the non-adversarial behavior of traffic participants \cite{predhumeau2022agent, Bonneaud2012ABD}, but to successfully and efficiently evaluate the performance of AVs in edge-case scenarios and identify potential weaknesses, it is crucial to incorporate the modeling of adversarial behavior. 
\subsection{Social Force Models}
The social force model is a widely studied method for modeling pedestrian behavior \cite{seyfried2006basics}. Researchers have built upon the foundational social force model by introducing new variations, such as the headed social force model \cite{bonneaud2012behavioral}, and improving the authenticity of the trajectories provided. However, these models have limitations including their efficiency and universality when applied to testing automated vehicles in edge-case scenarios or high-dimensional state spaces. 

\subsection{RL-based Methods}
The use of reinforcement learning (RL) for pedestrian behavior modeling has been studied widely. Researchers have employed RL frameworks for pedestrian agents to acquire navigation and obstacle avoidance behaviors \cite{9317723}, develop autonomous driving robotic vehicles in high-density pedestrian environments \cite{8202312}, and replicate human cognitive processes for anticipating hazards or obstacles \cite{app11125442}. The inverse RL method \cite{nasernejad2022multiagent} is also used to infer reward functions for pedestrian and vehicle interactions and collision avoidance mechanisms. 

\section{Method}
An edge-case scenario refers to an unusual or unforeseen situation \cite{kocc2021pedestrian} in automated vehicle systems that test the limits of the input space or operational design domain. These scenarios are difficult to anticipate during the AV system design process and also difficult to generate during testing, and therefore they may cause the system to make decisions outside of its normal parameters. To evaluate and test automated vehicles, it is important to simulate unexpected conditions to assess the robustness and safety of the algorithm.
\begin{figure}[!]
\centerline{\includegraphics[width=0.3\textwidth]{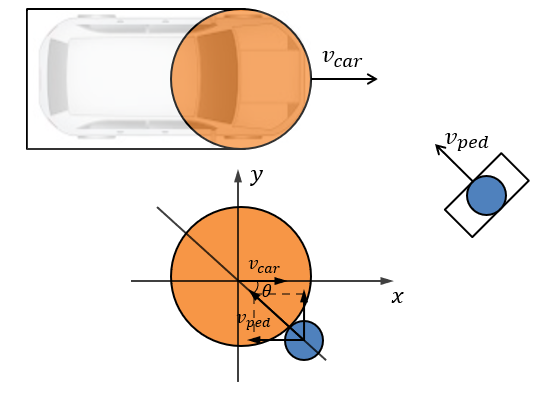}}
\caption{The kinematics and dynamics of pedestrian and vehicle}
\label{CMOMEN}
\vspace{-0.4cm}
\end{figure}

\subsection{Pedestrian collision problem}\label{AA}
Recent studies have mostly focused on path planning and collision avoidance, ant thus have ignored the use of collision lethality as a metric. Collision lethality is an important factor \cite{yastremska2022constitutes} to consider and can be measured using objective indicators such as relative velocity, kinetic energy, impulse, and change in velocity \cite{laureshyn2010evaluation}. We propose using collision momentum as a key factor in the design of RL methods for training pedestrian agents which may cause a lethal collision accident. Our experiments will consider Fig. \ref{ColPic} as a typical pedestrian-vehicle interaction scenario. 
\subsection{Pedestrian and vehicle kinematics and dynamics}\label{AA}
Fig. \ref{CMOMEN} shows that the pedestrian and vehicle's planar projections are extremely complex in the real-life scenario, and the maneuver simulation of the vehicle is related to the physics of wheel steering and vehicle dynamics. In our approach, we only consider the vehicle's front bumper and engine hood components as a circle (yellow) to simplify the model, because there are many frontal lethal vehicle-pedestrian collisions. We assume that the vehicle exhibits no lateral dynamics, no acceleration, and only the option of braking at $a=-2.5m/s^2$ to simulate the realistic emergency brake negative acceleration. Also, we consider the pedestrian as a discrete-time point mass model (blue) to simplify the model with the neural net dictating the $\Delta_{Heading}$ as the RL model's next timestep action. We only control the $\Delta_{Heading}$ at this stage of the work, rather than both heading and velocity, in order to reduce the number of neural net outputs. This seems reasonable, as the discrete timestep is small, 0.05s and the neural net can effectively generate a zero velocity by commanding the pedestrian to move back and forward in succession or any other sequence of heading commands that keep the pedestrian within a small area.

\subsection{Collision momentum formulation}\label{AA}
Fig. \ref{CMOMEN} shows the pedestrian and vehicle kinematic and dynamics model. The vehicle agent will react to the pedestrian's position and brake to decelerate as defined by the control algorithm under test. The collision process we used in this work is the perfect elastic collision momentum. The collision momentum change, which is a crucial indicator of the lethality of an accident, can be mathematically represented by the equation: $\Delta p = m ( v' - v)$.   The collision momentum change before and after the collision is:
\begin{equation}
\begin{aligned}
    \Delta p_{p} & = m_{p} (\sqrt{v_{px'}^2+v_{py'}^2}-v_p)\\
    v_{px'} & = \frac{\left(m_p-m_c\right)cos\theta v_{p}+2 m_c v_{c}}{m_c+m_p}-cos\theta v_{p}\\
    v_{py'} & = \frac{\left(m_p-m_c\right)sin\theta v_{p}}{m_c+m_p}-sin\theta v_{p}\\
\end{aligned}
\end{equation}
where $m_p$ and $m_c$ represent the mass of the pedestrian and vehicle respectively, $v_{p}$ and $v_{c}$ represent the velocity of the pedestrian and vehicle before the collision, and $v_{p'}$ represents the velocity of the pedestrian after the collision.

As previously noted, the magnitude of the collision momentum change is positively correlated with the severity of injury sustained by the pedestrian. Therefore, a larger change in momentum is indicative of a more damaging collision.

\begin{figure}[!]
\centerline{\includegraphics[width=0.42\textwidth]{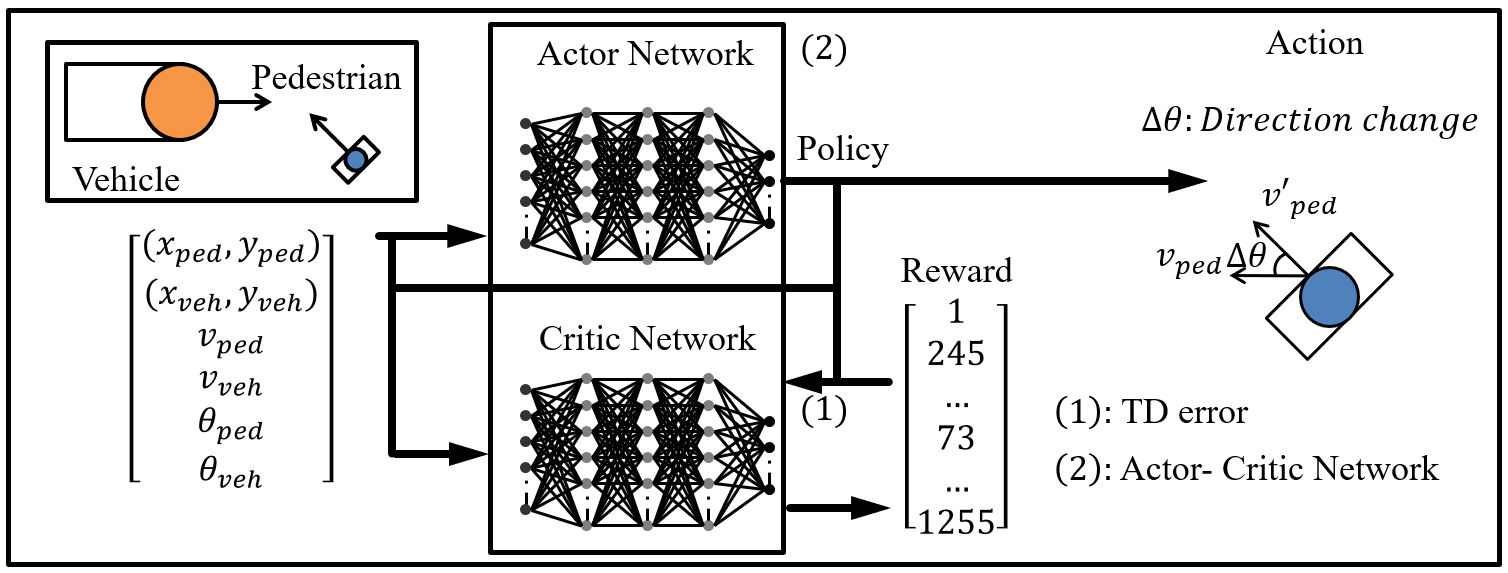}}
\caption{The DDPG-based DRL agent}
\label{Overviewdiagram}
\vspace{-0.4cm}
\end{figure}

\subsection{Markov decision process}\label{AA}
Our proposed pedestrian agent uses a small state, an action parameter, and a well-designed reward function. The simulation scenario can be designed as a Markov Decision Process (MDP) with the tuple $(S,A,P_a,r_a)$
 \subsubsection{$S$}
A set of states. The combination of the vehicle position $(x_{veh},y_{veh})$, pedestrian position $(x_{ped},y_{ped})$, vehicle velocity $v_{veh}$, pedestrian velocity $v_{ped}$, vehicle direction $\theta_{veh}$, and pedestrian direction $\theta_{ped}$. With the motion details of agents, we can calculate the collision momentum change of the pedestrian $\Delta p_{ped}$ and use it in the later reward design.
\subsubsection{$A$}
A set of actions. According to Fig. \ref{Overviewdiagram}, the action is the direction change $\Delta{\theta}$ of the next timestep. The pedestrian agent can use the action $\Delta{\theta}$ to interact with the environment and update the state $S$. 
\subsubsection{$P$}
The transition probability. $P_a(s,s')=Pr(s_{t+1}=s'|s_t=s,a_t=a)$. $P_a(s,s')$ is the probability from changing from state $s$ to state $s'$ when taking the action $a$.
\subsubsection{$r$}
Reward function $r(s_{t+1},s_t,a_t)$. The $r$ is the expected immediate reward of taking an exact action $a$ from state $s$ to state $s'$.

We want to find the policy function $\pi(s)$ which can output the best action for a particular state in order to maximize the expectation of cumulative future rewards:
\begin{equation}
    E\left[\sum_{t=0}^{\infty} \gamma^t r\left(s_t, s_{t+1},a_t\right)\right]
\end{equation}
where $\gamma$ is the discount factor and the range of $\gamma$ is $[0,1]$. A larger $\gamma$ dictates that the value of the future action takes a more important role in the cumulative reward function.

\subsection{Collision momentum based reward function}\label{AA}
In our proposed collision momentum model, there exist three possible state transitions shown in Fig. \ref{Rewardpicture}:
\begin{figure}[!]
\centerline{\includegraphics[width=0.4\textwidth]{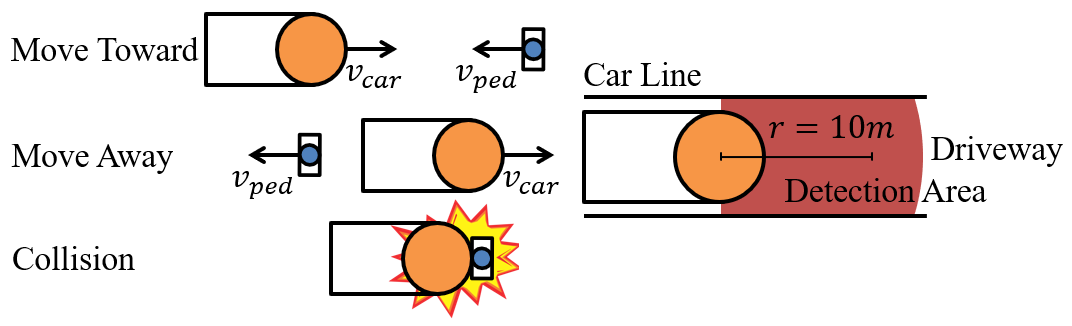}}
\caption{State transitions}
\label{Rewardpicture}
\vspace{-0.4cm}
\end{figure}
\begin{itemize}
    \item Pedestrian and Vehicle agents move toward each other. The reward for this timestep's transition is:
    \begin{equation}
        \frac{10}{1+\sqrt{\left| x_{ped}-x_{veh} \right|^2+\left| y_{ped}-y_{veh} \right|^2}}
    \end{equation}
    \item Pedestrian and Vehicle agents move away from each other. The reward of this timestep's transition is:
    \begin{equation}
        \frac{-10}{1+\sqrt{\left| x_{ped}-x_{veh} \right|^2+\left| y_{ped}-y_{veh} \right|^2}}-1
    \end{equation}
    \item Pedestrian and Vehicle agents collide with each other, with reward given by:
    \begin{equation}
        10(\frac{(m_p-m_c)v_{p}+2m_cv_c}{m_c+m_p}-v_p)m_p
    \end{equation}
\end{itemize}
    
 where $x_{ped}$ and $y_{ped}$ represent the position of the pedestrian, $x_{veh}$ and $y_{veh}$ represent the position of the vehicle, $v_p$ and $v_c$ represent the velocity of pedestrian and vehicle before the collision happens. The reward functions are shown in Table \ref{tab2}.
 
 In our proposed pedestrian agent RL model shown in Fig. \ref{Overviewdiagram} we use DDPG which is a model-free, off-policy reinforcement learning approach that trains agents to execute tasks with continuous action spaces to discern the optimal action for directional change at the subsequent timestep by utilizing the Reward function, an Actor-Critic network, an Experience Replay buffer, and normally distributed Gaussian Action Noise.

\section{Experiments}

\subsection{Social force based methodology}\label{AA}
As a baseline methodology for this study, we consider using the social force model to verify the pedestrian agent algorithm's efficiency in exploring the edge-case scenario and AV's weakness. The improved social force model formula is:
\begin{equation}
    \begin{aligned}
        F_{\alpha} &= F_v+F_d+F_p\\
        P_{\alpha}^{t+1} &= P_{\alpha}^t+v_{\alpha}^t\Delta t +\frac{F_{\alpha}^t\Delta t^2}{2 m_p}\\
        v_a^{t+1} &= v_{\alpha}^t+\frac{F_{\alpha}^t}{m_p}\Delta t\\
    \end{aligned}
    \label{eqSF}
\end{equation}
where $F_{\alpha}$ is the resultant force on the pedestrian, $F_{v}$ the force toward the moving vehicle, $F_{d}$ the force of crossing the street, and $F_{p}$ the velocity constraint force on pedestrian, $ P_{\alpha}^{t+1}$ and $P_{\alpha}^t$ are the positions of the pedestrian at time $t$ and $t+1$, and $\Delta t$ is the time step in each episode.

\begin{table}[!t]
\caption{DDPG Reward Design Table}
\begin{center}
 \resizebox{85mm}{18mm}{
\begin{tabular}{|c|c|c|c|}
\hline
\textbf{}&\multicolumn{3}{|c|}{\textbf{Reward Design}} \\
\hline
\thead{\textbf{State} \\\textbf{transition}} & \thead{Social \\Force\\ Method$^{1}$} & \thead{RL \\Baseline\\ Method$^{2}$} & \thead{Collision Momentum\\ RL Method \textbf{(proposed)}}\\
\hline
\thead{Move\\ Away} & - & $+1$ & $\frac{10}{1+\sqrt{| x_{ped}-x_{veh} |^{2}+| y_{ped}-y_{veh}|^{2}}}$ \\
\hline
\thead{Move\\ Toward} & - & $-2$ & $\frac{-10}{1+\sqrt{| x_{ped}-x_{veh}|^{2}+|y_{ped}-y_{veh}|^{2}}}-1$ \\
\hline
Collision & - & $3000$ & $10(\frac{(m_p-m_c)v_{p}+2m_cv_c}{m_c+m_p}-v_p)m_p$\\  
\hline
\multicolumn{4}{l}{$^{\mathrm{1}}$ Social force model proposed by\cite{yang2018social}  $^{\mathrm{2}}$ RL baseline model proposed by\cite{9931272}} 
\end{tabular}}
\label{tab2}
\end{center}
\vspace{-0.6cm}
\end{table}

\subsection{RL methodology}\label{AA}
We proposed two reinforcement methodologies for pedestrian agents which have different reward function designs. 
We study the following combination of well-trained pedestrian models and vehicle driving algorithms:
\begin{itemize}
    \item The vehicle agent with constant desired velocity and the pedestrian agent with deep deterministic policy gradient and a reward function involving collision signal and direction.
    \item The vehicle agent with constant desired velocity and braking capability and the pedestrian agent with deep deterministic policy gradient with a reward function involving a collision signal and direction.
    \item The vehicle agent with constant desire velocity and braking and the pedestrian agent with deep deterministic policy gradient with a reward function involving collision momentum, direction, and distance.
\end{itemize}

\subsection{Implementation}\label{AA}
Like the other RL methods used in modeling pedestrian agents, the main process of training pedestrian agents is making sure the pedestrian is able to explore the edge-case scenario and the neural net can learn from the collision edge-case scenario. Thus, the main point of the scenario design is to ensure that during the exploration, a significant portion of the episodes are useful for the pedestrian agent to maximize the reward and make the collision more lethal. 
Therefore, we proposed the training method in Python 3.6.10 based on the RL implementation framework that is Stable-Baseline 3 \cite{raffin2021stable}. And we modified the RL environment based on the Gym environment "Pendulum-v1" \cite{brockman2016openai}. 

\paragraph{Scenario Parameters}
The initial $v_p$ and $v_c$ of each episode are $2m/s$ and $7m/s$, and the initial position of pedestrian $(x_{ped},y_{ped})$ and vehicle $(x_{veh},y_{veh})$ are $(50m,-5m)$ and $(0m,0m)$ respectively. In trying to explore the randomness and enrich the training data set variety, we limit the pedestrian initial position to $(50\pm10m,-5m)$ which means the pedestrian will randomly choose the x-axis initial position from $40m$ to $60m$ for each training episode.

\begin{algorithm}[!t]
	\caption{DDPG Pedestrian Agent Episode Algorithm} 
	\label{alg3} 
	\begin{algorithmic}[1]
		\REQUIRE The pedestrian and vehicle position $P(x_p,y_p)$ and $V(x_v,y_v)$, sidewalk area $S_{sd}$, driveway area $S_{dw}$, and vehicle brake acceleration $a_v$.
		\STATE Start episode        
		\WHILE{$(x_p,y_p)\in S_{dw}$} 
        \IF{$|PV|<10m$}
        \STATE $a_v = -3.5 m/s$ 
        \IF{Collision happened}
        \STATE End current training episode. 
        \ELSIF{No Collision happened}
        \STATE Interact with the environment. Turn to the current episode's next time step. Return to $Step_{(2)}$
        \ENDIF
        \ELSIF{$|PV|>10m$}
        \STATE Interact with the environment. Turn to the current episode's next time step. Return to $Step_{(2)}$
        \ENDIF
        \ENDWHILE
        \WHILE{$(x_p,y_p)\in S_{sw}$}
        \STATE $a_v = 0 m/s$
        \STATE Interact with the environment. Turn to the current episode's next time step. Return to $Step_{(2)}$
		\ENDWHILE
	\end{algorithmic} 
\end{algorithm}

\paragraph{DDPG architecture and hyperparameters}
As shown in Fig. \ref{Overviewdiagram} we establish the actor-critic neural network as four fully connected networks. Each network has three hidden layers and uses the array [state dimension, 512, 256, action dimension] as the network node. In the DDPG algorithm, we also used the soft-update network method to update the network from the target network.
The soft-update parameter $\tau$ is 0.005. The memory capacity and the batch size of the experience replay buffer in the DDPG algorithm are 10000 and 1000, $\gamma$ in the bellman optimality equation is 0.9, and the learning rate of the actor network and critic network are 0.001 and 0.002.
\vspace*{-0.15cm}
\section{Results}
We implemented the social force based method, RL baseline method, and our proposed collision momentum RL method. The trajectories of the pedestrians for each case are shown in Fig. \ref{AL2}.
\subsection{Social force based method}\label{AA}
The crossing street force of the designed social force models that are shown in \eqref{eqSF} is necessary. We implemented two different social force models and there will be no collision happening on the model without the crossing street force in our designed scenario. The social force model with the crossing street force will act well and reach a high collision momentum after specific value selection. The trajectory of the social force method agent interacts with the vehicle which will brake in the emergency situation is shown in Fig. \ref{AL2}a. 
\subsection{RL Baseline method}\label{AA}
The two trajectories of the RL baseline method agent \cite{9931272}, interacting with the vehicle agent which either will or will not brake in an emergency situation, perform similarly and the latter one will reach higher collision momentum and cause bigger lethality. But following the purpose of simulating the more realistic scenario, we take the former vehicle algorithm which is able to brake. The trajectory of the RL baseline method agent interacts with the vehicle which will brake in the emergency situation is shown in Fig. \ref{AL2}b, where we can clearly find that the pedestrian agent tries to stay in the sidewalk area and not trigger the AV's emergency brake mode until the AV is close enough to generate the potential collision. 
        
\subsection{Proposed collision momentum RL method}\label{AA}
\begin{figure}[tbp]
        \centerline{\includegraphics[width=0.45\textwidth]{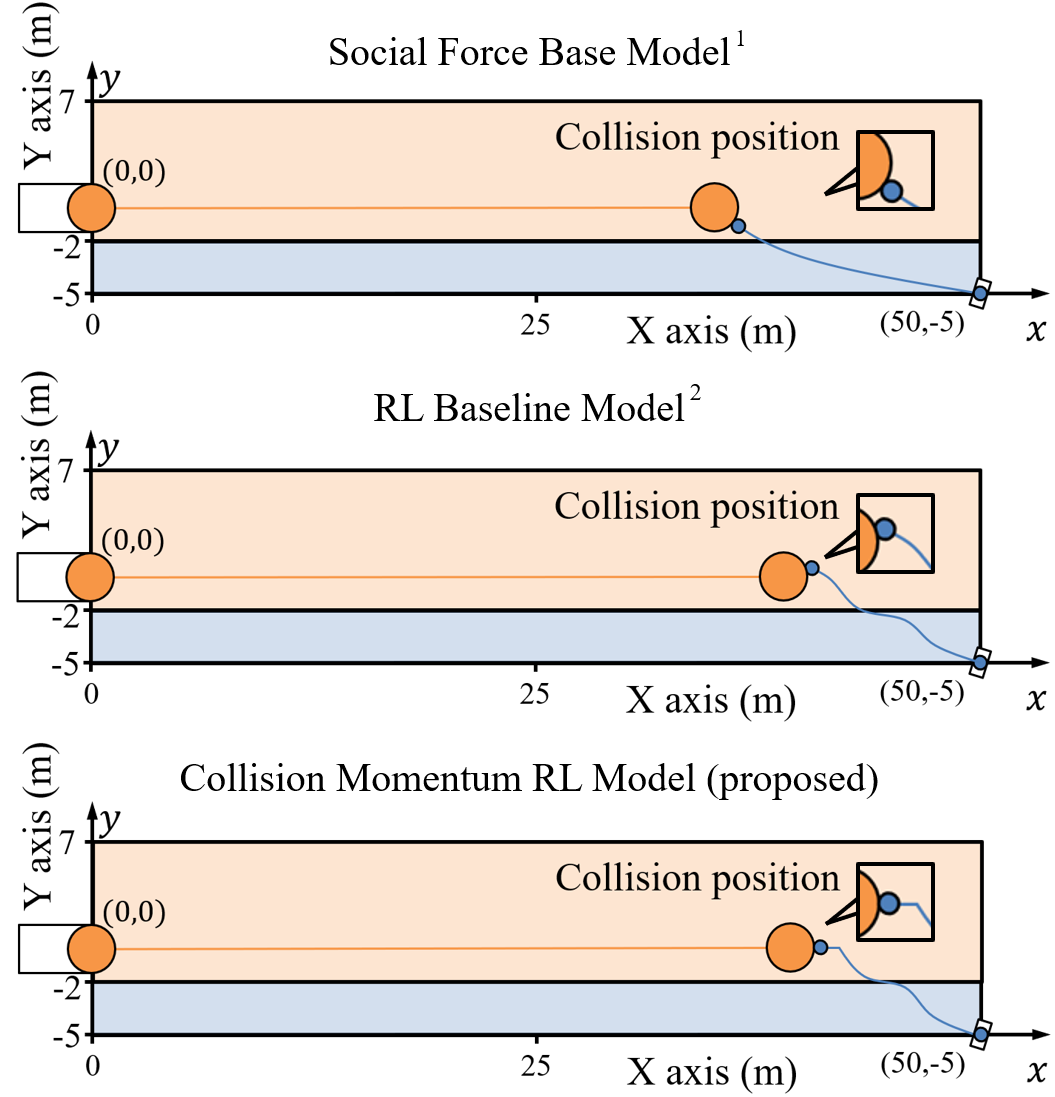}}
        \caption{Pedestrian Trajectory of Collision Accident, $^{\mathrm{1}}$ Social force model proposed by\cite{yang2018social},  $^{\mathrm{2}}$ RL baseline model proposed by\cite{9931272}. }
        \label{AL2}
        \vspace{-0.5cm}
        \end{figure}

Our proposed collision momentum RL method which considers the collision momentum in the reward design process also interacts with the vehicle agent which will brake in an emergency situation. The trajectory of the collision momentum RL method is shown in Fig. \ref{AL2}c. 
The trajectory of the earlier time steps with the pedestrian agent is extremely similar to the RL baseline method's trajectory, but in the last few time steps, the pedestrian agent learns the best action to maximize the collision momentum to maximize and the lethality. Thus, the pedestrian agent will move directly toward the vehicle agent, that is to the left for a head-on collision. 
\subsection{Reward function}\label{AA}
\begin{figure}[!]
\centering
\includegraphics[width=0.5\textwidth]{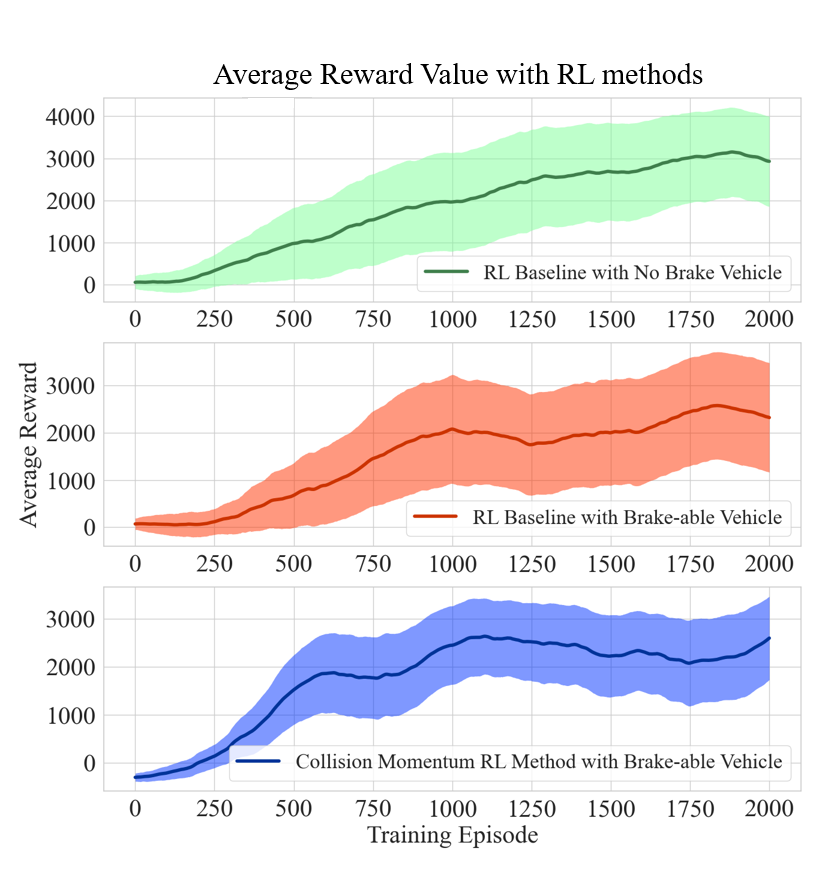}
\caption{Average episode reward with confidence intervals of eight experimental runs. Each run was initialized with random neural network weights. $^{\mathrm{1}}$ RL baseline model proposed by\cite{9931272}, $^{\mathrm{2}}$ RL method proposed by us.}
\label{RF}
\vspace{-0.4cm}
\end{figure}

We trained the pedestrian agent with different pseudo-random numbers in the Python RL framework.  The average reward function for each of the three different scenarios is shown in Fig. \ref{RF}.

A comparison between the average reward value of the RL baseline and our proposed method is not informative because of the different design methods of the reward functions. Compared to the RL Baseline Methods in non-brake and brake-capable vehicle scenarios, the former scenario reward function will reach a higher reward because of the simpler vehicle algorithm. Scenario 2 and scenario 3 have a reward function decrease at the episode range of $1000$ to $1250$ and range of $600$ to $800$, respectively. At the beginning of the decrease, the agent has already learned the general direction toward the moving vehicle. But with the closer distance between the vehicle and the pedestrian, the vehicle starts to brake with a constant deceleration and the agent needs to learn the updated performance of the braking vehicle, and during the learning process, the vehicle will also react to the more intelligent agent. Finally, the pedestrian will be well-trained before the network is over-fit. Also, it's clearly shown that our proposed RL method converges faster than the RL baseline methods.
\subsection{Well-trained model recall and verification}\label{AA}
With well-trained pedestrian agents, we can recall the Actor-Critic neural network to test the efficiency in exploring the edge-case scenario and AV's weakness by using the value of the collision momentum change as the judgment factor. We test each algorithm in the stochastic start scenario, in which the pedestrian has a stochastic start position within a rectangle area ranging from $40$ to $60$ on the x-axis and from $-3$ to $-6$ on the y-axis, conclude the well-trained algorithm's collision momentum change of the pedestrian agent with different start positions and then calculate the average collision momentum change and the variance of the data. The experiment recall and calculation results are shown in Table \ref{tab3}.
\begin{table}[!]
\caption{Average Collision Momentum Change}
\begin{center}
\scalebox{0.9}{
\begin{tabular}{|c|c|c|c|}
\hline
\thead{\textbf{Algorithm}} & \thead{Social Force\\ Based Method$^{1}$} & \thead{RL Baseline\\ Method$^{2}$} & \thead{\textbf{Collision}\\ \textbf{Momentum}\\ \textbf{RL Method}\\ \textbf{(Proposed)}}\\ 
\hline
\thead{\textbf{Collision}\\\textbf{Momentum}\\ \bm{$(kg*m/s)$}} & \thead{$376.49$\\ $\pm105.95$} & \thead{$285.71$\\ $\pm0.142$} & \thead{\bm{$473.78$}\\ \bm{$\pm15.39$}}\\ 
\hline
\multicolumn{4}{l}{$^{\mathrm{1}}$Social force model proposed by\cite{yang2018social}  $^{\mathrm{2}}$RL baseline model proposed by\cite{9931272}}
\end{tabular}}
\label{tab3}
\end{center}
\end{table}

\section{Conclusion}
In this work, we propose a new DRL approach to training the adversarial pedestrian for exploring the weakness of the AV. The virtual environment implementation and verification show that: the proposed method is able to learn to exploit the AV's weaknesses, the reinforcement learning algorithm is more lethal to the social force based pedestrian motion algorithm, and the DDPG algorithm designed by collision momentum is more lethal to pedestrians than that designed by collision impact signal (RL baseline method). Finally, this work can prove the efficiency and effect of the proposed method, and future work will focus on improving the RL algorithm based on more complicated AV algorithms and testing the commercial vehicle algorithms.

\bibliographystyle{IEEEtran}
\bibliography{IEEEabrv,references}

\begin{thebibliography}{10}
\providecommand{\url}[1]{#1}
\csname url@samestyle\endcsname
\providecommand{\newblock}{\relax}
\providecommand{\bibinfo}[2]{#2}
\providecommand{\BIBentrySTDinterwordspacing}{\spaceskip=0pt\relax}
\providecommand{\BIBentryALTinterwordstretchfactor}{4}
\providecommand{\BIBentryALTinterwordspacing}{\spaceskip=\fontdimen2\font plus
\BIBentryALTinterwordstretchfactor\fontdimen3\font minus
  \fontdimen4\font\relax}
\providecommand{\BIBforeignlanguage}[2]{{%
\expandafter\ifx\csname l@#1\endcsname\relax
\typeout{** WARNING: IEEEtran.bst: No hyphenation pattern has been}%
\typeout{** loaded for the language `#1'. Using the pattern for}%
\typeout{** the default language instead.}%
\else
\language=\csname l@#1\endcsname
\fi
#2}}
\providecommand{\BIBdecl}{\relax}
\BIBdecl

\bibitem{chan2017advancements}
C.-Y. Chan, ``Advancements, prospects, and impacts of automated driving
  systems,'' \emph{International Journal of Transportation Science and
  Technology}, vol.~6, no.~3, pp. 208--216, 2017.

\bibitem{bengler2014three}
K.~Bengler, K.~Dietmayer, B.~Farber, M.~Maurer, C.~Stiller, and H.~Winner,
  ``Three decades of driver assistance systems: Review and future
  perspectives,'' \emph{IEEE Intelligent Transportation Systems Magazine},
  vol.~6, no.~4, pp. 6--22, 2014.

\bibitem{bar2014recent}
A.~Bar~Hillel, R.~Lerner, D.~Levi, and G.~Raz, ``Recent progress in road and
  lane detection: a survey,'' \emph{Machine Vision and Applications}, vol.~25,
  no.~3, pp. 727--745, 2014.

\bibitem{bonnefon2016social}
J.-F. Bonnefon, A.~Shariff, and I.~Rahwan, ``The social dilemma of autonomous
  vehicles,'' \emph{Science}, vol. 352, no. 6293, pp. 1573--1576, 2016.

\bibitem{beringhoff2022thirty}
F.~Beringhoff, J.~Greenyer, C.~Roesener, and M.~Tichy, ``Thirty-one challenges
  in testing automated vehicles: Interviews with experts from industry and
  research,'' in \emph{2022 IEEE Intelligent Vehicles Symposium (IV)}, 2022,
  pp. 360--366.

\bibitem{helle2016testing}
P.~Helle, W.~Schamai, and C.~Strobel, ``Testing of autonomous
  systems--challenges and current state-of-the-art,'' in \emph{INCOSE
  International Symposium}, vol.~26, no.~1, 2016, pp. 571--584.

\bibitem{mullins2017automated}
G.~E. Mullins, P.~G. Stankiewicz, and S.~K. Gupta, ``Automated generation of
  diverse and challenging scenarios for test and evaluation of autonomous
  vehicles,'' in \emph{2017 IEEE International Conference on Robotics and
  Automation (ICRA)}, 2017, pp. 1443--1450.

\bibitem{yang2018social}
D.~Yang, {\"U}.~{\"O}zg{\"u}ner, and K.~Redmill, ``Social force based
  microscopic modeling of vehicle-crowd interaction,'' in \emph{2018 IEEE
  Intelligent Vehicles Symposium (IV)}, 2018, pp. 1537--1542.

\bibitem{predhumeau2022agent}
M.~Pr{\'e}dhumeau, L.~Mancheva, J.~Dugdale, and A.~Spalanzani, ``Agent-based
  modeling for predicting pedestrian trajectories around an autonomous
  vehicle,'' \emph{Journal of Artificial Intelligence Research}, vol.~73, pp.
  1385--1433, 2022.

\bibitem{Bonneaud2012ABD}
S.~Bonneaud and W.~H. Warren, ``A behavioral dynamics approach to modeling
  realistic pedestrian behavior,'' 2012.

\bibitem{chen2017socially}
Y.~F. Chen, M.~Everett, M.~Liu, and J.~P. How, ``Socially aware motion planning
  with deep reinforcement learning,'' in \emph{2017 IEEE/RSJ International
  Conference on Intelligent Robots and Systems (IROS)}, 2017, pp. 1343--1350.

\bibitem{noauthor_pedestrian_2022}
{National Highway Traffic Safety Administration},
  ``\BIBforeignlanguage{en}{Traffic safety facts 2019 data: Pedestrians},''
  \url{https://crashstats.nhtsa.dot.gov/Api/Public/Publication/813079external},
  May 2022.

\bibitem{noauthor_reported_nodate}
\BIBentryALTinterwordspacing
``\BIBforeignlanguage{en}{Reported road casualties {Great} {Britain}:
  pedestrian factsheet 2021}.'' [Online]. Available:
  \url{https://www.gov.uk/government/statistics/reported-road-casualties-great-britain-pedestrian-factsheet-2021/reported-road-casualties-great-britain-pedestrian-factsheet-2021}
\BIBentrySTDinterwordspacing

\bibitem{wang2020safety}
J.~Wang, L.~Zhang, Y.~Huang, and J.~Zhao, ``Safety of autonomous vehicles,''
  \emph{Journal of Advanced Transportation}, vol. 2020, 2020.

\bibitem{liu2019edge}
S.~Liu, L.~Liu, J.~Tang, B.~Yu, Y.~Wang, and W.~Shi, ``Edge computing for
  autonomous driving: Opportunities and challenges,'' \emph{Proceedings of the
  IEEE}, vol. 107, no.~8, pp. 1697--1716, 2019.

\bibitem{ni2011modelling}
Y.~Ni and K.~Li, ``Modelling pedestrian behavior at signalized intersections: A
  case study in shanghai,'' in \emph{ICTIS 2011: Multimodal Approach to
  Sustained Transportation System Development: Information, Technology,
  Implementation}, 2011, pp. 1745--1754.

\bibitem{martinez2017modeling}
F.~Martinez-Gil, M.~Lozano, I.~Garc{\'\i}a-Fern{\'a}ndez, and F.~Fern{\'a}ndez,
  ``Modeling, evaluation, and scale on artificial pedestrians: a literature
  review,'' \emph{ACM Computing Surveys (CSUR)}, vol.~50, no.~5, pp. 1--35,
  2017.

\bibitem{moussaid2010walking}
M.~Moussa{\"\i}d, N.~Perozo, S.~Garnier, D.~Helbing, and G.~Theraulaz, ``The
  walking behaviour of pedestrian social groups and its impact on crowd
  dynamics,'' \emph{PloS One}, vol.~5, no.~4, p. e10047, 2010.

\bibitem{kielar2020artificial}
P.~Kielar and A.~Borrmann, ``An artificial neural network framework for
  pedestrian walking behavior modeling and simulation,'' \emph{Collective
  Dynamics}, vol.~5, pp. 290--298, 2020.

\bibitem{helbing1995social}
D.~Helbing and P.~Molnar, ``Social force model for pedestrian dynamics,''
  \emph{Physical Review E}, vol.~51, no.~5, p. 4282, 1995.

\bibitem{bonneaud2012behavioral}
S.~Bonneaud and W.~H. Warren, ``A behavioral dynamics approach to modeling
  realistic pedestrian behavior,'' in \emph{6th International Conference on
  Pedestrian and Evacuation Dynamics}, 2012, pp. 1--14.

\bibitem{kiran2021deep}
B.~R. Kiran, I.~Sobh, V.~Talpaert, P.~Mannion, A.~A. Al~Sallab, S.~Yogamani,
  and P.~P{\'e}rez, ``Deep reinforcement learning for autonomous driving: A
  survey,'' \emph{IEEE Transactions on Intelligent Transportation Systems},
  2021.

\bibitem{seyfried2006basics}
A.~Seyfried, B.~Steffen, and T.~Lippert, ``Basics of modelling the pedestrian
  flow,'' \emph{Physica A: Statistical Mechanics and its Applications}, vol.
  368, no.~1, pp. 232--238, 2006.

\bibitem{9317723}
M.~Everett, Y.~F. Chen, and J.~P. How, ``Collision avoidance in pedestrian-rich
  environments with deep reinforcement learning,'' \emph{IEEE Access}, vol.~9,
  pp. 10\,357--10\,377, 2021.

\bibitem{8202312}
Y.~F. Chen, M.~Everett, M.~Liu, and J.~P. How, ``Socially aware motion planning
  with deep reinforcement learning,'' in \emph{2017 IEEE/RSJ International
  Conference on Intelligent Robots and Systems (IROS)}, 2017, pp. 1343--1350.

\bibitem{app11125442}
\BIBentryALTinterwordspacing
T.-T. Trinh and M.~Kimura, ``The impact of obstacle’s risk in pedestrian
  agent’s local path-planning,'' \emph{Applied Sciences}, vol.~11, no.~12,
  2021. [Online]. Available: \url{https://www.mdpi.com/2076-3417/11/12/5442}
\BIBentrySTDinterwordspacing

\bibitem{nasernejad2022multiagent}
P.~Nasernejad, T.~Sayed, and R.~Alsaleh, ``Multiagent modeling of
  pedestrian-vehicle conflicts using adversarial inverse reinforcement
  learning,'' \emph{Transportmetrica A: Transport Science}, pp. 1--35, 2022.

\bibitem{kocc2021pedestrian}
M.~Ko{\c{c}}, E.~Yurtsever, K.~Redmill, and {\"U}.~{\"O}zg{\"u}ner,
  ``Pedestrian emergence estimation and occlusion-aware risk assessment for
  urban autonomous driving,'' in \emph{2021 IEEE International Intelligent
  Transportation Systems Conference (ITSC)}, 2021, pp. 292--297.

\bibitem{yastremska2022constitutes}
O.~Yastremska-Kravchenko, A.~Laureshyn, C.~D'Agostino, and A.~Varhelyi, ``What
  constitutes traffic event severity in terms of human danger perception?''
  \emph{Transportation Research Part F: Traffic Psychology and Behaviour},
  vol.~90, pp. 22--34, 2022.

\bibitem{laureshyn2010evaluation}
A.~Laureshyn, {\AA}.~Svensson, and C.~Hyd{\'e}n, ``Evaluation of traffic
  safety, based on micro-level behavioural data: Theoretical framework and
  first implementation,'' \emph{Accident Analysis \& Prevention}, vol.~42,
  no.~6, pp. 1637--1646, 2010.

\bibitem{9931272}
P.~Phueakthong, J.~Varagul, and N.~Pinrath, ``Deep reinforcement learning based
  mobile robot navigation in unknown environment with continuous action
  space,'' in \emph{2022 5th International Conference on Intelligent Autonomous
  Systems (ICoIAS)}, 2022, pp. 154--158.

\bibitem{raffin2021stable}
A.~Raffin, A.~Hill, A.~Gleave, A.~Kanervisto, M.~Ernestus, and N.~Dormann,
  ``Stable-baselines3: Reliable reinforcement learning implementations,''
  \emph{The Journal of Machine Learning Research}, vol.~22, no.~1, pp.
  12\,348--12\,355, 2021.

\bibitem{brockman2016openai}
G.~Brockman, V.~Cheung, L.~Pettersson, J.~Schneider, J.~Schulman, J.~Tang, and
  W.~Zaremba, ``Openai gym,'' \emph{arXiv preprint arXiv:1606.01540}, 2016.

\end{thebibliography}
\end{document}